\documentclass[a4paper, 10pt, conference]{ieeeconf}      

\IEEEoverridecommandlockouts                              

\usepackage{graphicx}

\title{\LARGE \bf
``Who is Driving around Me?”  \\ Unique Vehicle Instance Classification using Deep Neural Features.}

\author{Tim Oosterhuis$^{1}$ and Lambert Schomaker$^{2}$
\thanks{*This work was not supported by any organization}
\thanks{$^{1}$Tim Oosterhuis is a MSc graduate in Artificial Intelligence and Computing Science, 
        University of Groningen, 9700 AB Groningen, The Netherlands
        {\tt\small t.s.oosterhuis@rug.nl}}%
\thanks{$^{2}$Prof. Dr. Lambert Schomaker is professor in Artificial Intelligence, University of Groningen,
        9700 AB Groningen, The Netherlands
        {\tt\small l.r.b.schomaker@rug.nl}}%
}

\begin{document}

\maketitle
\thispagestyle{empty}
\pagestyle{empty}

\begin{abstract}

Awareness of other traffic is needed for self-driving cars to operate in the real world. Currently, multi-vehicle tracking is hindered by ID-switches. Accurate identification would solve this. In this paper, we show how the intrinsic feature maps of an object detection CNN can be used to uniquely identify vehicles from a dash-cam feed at real-time speed. Feature maps of a pretrained `YOLO' network are used to create 700 deep integrated feature signatures from 20 different images of 35 vehicles from a high resolution dataset and 340 signatures from 20 different images of 17 vehicles of a lower resolution tracking benchmark dataset. The YOLO network was trained to classify general object categories, e.g. classify a detected object as a `car' or `truck'. 5-Fold nearest neighbor (1NN) classification was used on DIFS created from feature maps in the middle layers of the network to correctly identify unique vehicles at a rate of 96.7\% for the high resolution data and with a rate of 86.8\% for the lower resolution data. We conclude that a deep neural detection network trained to distinguish between different classes can be successfully used to identify different instances belonging to the same class, using deep integrated feature signatures.
\end{abstract}

\section{INTRODUCTION}
\label{intro}

In recent years, developments within artificial intelligence and machine learning have made it possible for deep convolutional neural networks (CNNs) to detect and classify objects in a larger scene image at near real-time speeds. Among many other possible applications, this would allow for the detection of traffic participants from a vehicle-mounted camera, which would be a vital step in realizing the interaction of autonomous vehicles in real-world traffic. \\
However, for a potential autonomous vehicle to be truly intelligent, it has to be situationally aware. Frame-by-frame detection of surrounding traffic is needed, but it is not enough. Although state-of-the-art object detection CNNs attain very high average accuracy, they are not perfect. Due to the inherent nature of the complexity involved in training deep neural networks, they can generate unpredictable outcomes when detecting objects on new data. For dash-cam video streams of traffic this can include momentary spurious detections, momentary misses of vehicles in plain sight and partial detections where two vehicles are detected as a single vehicle, or only part of a vehicle is detected \cite{Oosterhuis2019who}.  Additionally, the traffic environment can be chaotic due to the large number of participants and frequent moments of temporary occlusion, e.g. when vehicles overtake each other. In order to compensate for all of this, an aspect of cognition is required in order to predict future positions of other traffic participants proactively and prevent any possible collisions. \\
Multiple different tracking approaches have been used to track the trajectory of vehicles and predict their future positions using the bounding boxes obtained from an object detection CNN over a number of frames. These include traditional mathematical models like the extended Kalman filter \cite{melo2004viewpoint} as well as more recently developed methods such as specialized correlation filters \cite{danelljan2016beyond}, aggregated local flow descriptors \cite{choi2015near}, a Poisson multi-Bernoulli mixture filter \cite{scheidegger2018mono}. \\
However such tracking models by default are `blind' to the content of these bounding boxes, which is an unique vehicle which has the same uniquely identifiable characteristics in each frame. This can lead to unnecessary ID-switches and other tracking problems stemming from occlusion and crowded scenes \cite{zamir2012global}. \\
In this paper, we present a means of highly accurate, fast unique vehicle instance classification through the creation of Deep Integrated Feature Signatures (DIFS), which can be embedded in an online multi-vehicle detector and tracker, thereby aiding the potential of situational awareness with regard to specific other traffic participants from the perspective of a self-driving vehicle. 

\section{RELATED WORK}
\label{related}
In addition to the vehicle detection and tracking methods described in Section \ref{intro}, deep neural networks have previously been applied to solve a number of problems in the specific domain of scene understanding in autonomous driving. Recurrent neural networks receiving cues from  dash-cam footage have been deployed for accident detection  \cite{chan2016anticipating}, and for intention and path prediction of cyclists \cite{pool2019context} and cars \cite{torstensson2019using}. Although the research in this paper is focused on input from a single camera, multi-sensor solutions for detection and tracking of road users in 3D using CNNs have also been developed using for example additional input from stereo-camera \cite{frossard2018end} or LIDAR \cite{roth2019deep}. \\ 
The YOLO object detection network which is used in this study has previously been adapted for pedestrian detection for video surveillance purposes \cite{molchanov2017pedestrian}. The YOLO network features have also been used to train a deep CNN to measure similarity of pedestrian detections in the same frame for the purpose of crowd counting \cite{yang2019pedestrian}. The possibility of pedestrian detection using deep learning methods for the purpose of autonomous driving has also been studied. Deep convolutional neural networks have been used to detect pedestrians on crosswalks \cite{guidolini2018handling} and on highway entrances \cite{he2017real}.

\section{METHODS}
\label{methods}

\subsection{Vehicle Detection}

\begin{figure}[b]
\centering
\includegraphics[width=\linewidth]{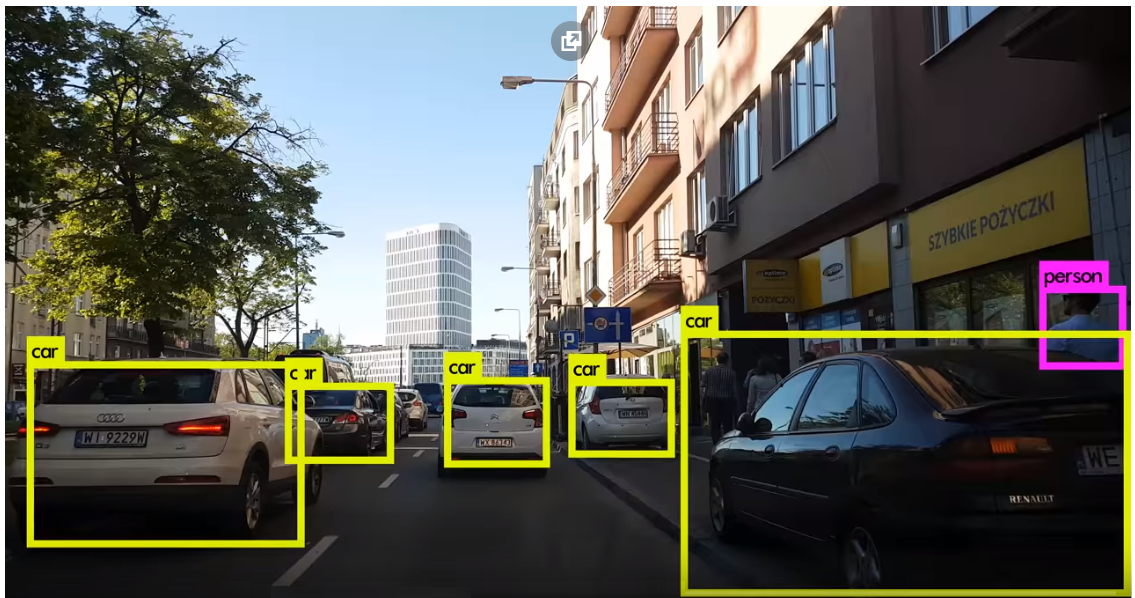}
\caption[YOLO example]{Traffic detection from a dash-cam feed using the YOLO v3 deep neural object detection network.}
\label{fig:yolo_output}
\end{figure}

\begin{figure}[t]
\centering
\includegraphics[width=\linewidth]{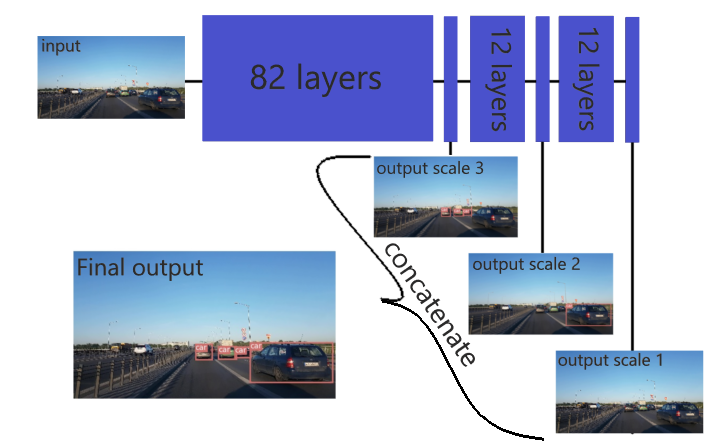}
\caption[YOLO architecture]{An abstract diagram of the architecture of the YOLO v3 network, showing the detection of vehicles at multiple scales starting from an input image.}
\label{fig:yolo_arch}
\end{figure}

For the vehicle detection step a pretrained YOLO v3 \cite{redmon2018yolov3} was used. YOLO v3 is a deep convolutional neural network trained to detect and classify objects within a larger scene. The YOLO v3 network outputs class labels, a confidence score and bounding boxes for each detected object in an input image. Fig. \ref{fig:yolo_output} shows an example of the output of YOLO vehicle detection in a traffic situation. The YOLO v3 network used in this research was pretrained on the COCO \cite{lin2014microsoft} dataset, which consists of labeled scene object data in 80 general object classes, including four vehicle classes (car, truck, bus and motorcycle). \\
YOLO v3 is multi-scale with three output layers for different sizes of boundary boxes and 106 layers total (see Fig. \ref{fig:yolo_arch}), all of which are convolutional. The YOLO v3 network also uses two kinds of skip connections. Firstly, throughout the network there are short skip connections named routes, in which three layers are skipped to form small residual blocks. Secondly, there are two long skip connections named shortcuts, where the output from early layers in the network is concatenated to residual layers just before the two last output layers. Upsampling layers are used to make this concatenation of early layer output via skip connections possible. \\

For this research a deliberate choice was made to use a pretrained deep neural network, because real-life traffic is a very dynamic environment, where the surrounding vehicles that need to be identified vary from moment to moment, which would never be truly reflected in a training set consisting of only a select number of vehicles. Furthermore, successfully training very deep neural networks is a time-intensive affair, due to all the manual work it entails in the form of manual labeling, data preprocessing and augmentation, network configuration and hyper-parameter tuning.\\ 
The YOLO architecture was chosen for this research over other deep neural object detection architectures such as RESNET or Inception, for two reasons. Firstly, YOLO is able to detect objects at near real-time speed on conventional hardware, while retaining a high level of accuracy \cite{redmon2018yolov3}.  We felt that the near real-time speed provided by the YOLO architecture would be desirable for any multi-vehicle tracking solution that could potentially have a practical application. An additional advantage of YOLO is that its performance has been demonstrated to be generalizable. Previously, the YOLO network has been used successfully to detect and classify objects in paintings, after being trained on photo-realistic examples \cite{redmon2016you}. A choice for YOLO v3 was made over earlier versions of YOLO, because YOLO v3 was reported to be more accurate than previous versions, especially when detecting small objects due to it's multi-scale output layers \cite{redmon2018yolov3}. We expected this feature to be useful when detecting vehicles from far away. \\
 . 

\subsection{Vehicle Instance Classification}

The YOLO v3 network reduces the dimensionality of dash-cam video data by several orders of magnitude from more than a million pixels to a list of detected traffic participants per frame. Despite this, the output from the network is still subject to a big data problem when considering the trajectories of unique vehicle instances from frame to frame, because of the exponential number of ways different vehicle detections could be combined into multi-frame trajectories.  The number of different unique vehicle instances in traffic is nearly infinite. The classical multi-class-based approach with a fixed number of classes is therefore not an option.\\ 
One potential identification method which may seem promising, is automatic licence plate recognition (ALPR). ALPR is an established technique with wide usage in the fields of police work and security. However, the typical ALPR application involves identifying a specific vehicle in one single frame to establish its presence in a general area around a general time. Being able to (re)identify vehicles at any moment these are in view of a vehicle-mounted camera, requires a flexibility which we found automatic licence plate recognition does not provide, because a vehicle's licence plate is either not visible or not readable during the majority of its trajectory on camera due to distance, angle, occlusion, poor illumination or motion blur. In addition to this, a licence plate recognition system, even if somehow perfect, would not be able to generalize to bicycles and other traffic participants without licence plates. \\

\subsubsection{Signatures based on Feature Representations}
The classical multi-class-based approach with a fixed number of classes is not an option for accurate and generalizable instance classification of unique vehicles. However, an identifying signature for each detected vehicle instance based on a feature representation of the vehicle can provide a reliable means of instance classification through similarity. \\
Such a feature-based signature could be created in multiple ways. A traditional approach would be extracting fixed features from the detected vehicle image algorithmically, such as with a Scale-Invariant Feature Transform (SIFT) \cite{lowe1999object} or by creating color histograms \cite{choi2015near} \cite{zamir2012global}. However, this requires an extra processing step, and preliminary research has shown that the accuracy of these methods is low for the complex problem of classifying unique vehicle instances in real traffic conditions from frames of dash-cam footage \cite{Oosterhuis2019who}. 
\subsubsection{Deep Feature Signatures}
In this research, the possibility of generating a feature-based signature automatically is explored. This is achieved by using a CNN as a deep feature extractor. The first layers of a typical convolutional network trained for image classification contain low-level feature representations, such as edges, corners and color patterns. Later layers contain more complex abstract feature maps, which finally culminate into single values relating the input image to each learned class in the last layer The phenomenon of how low-level features are being combined into more complex ones within a CNN, is also the guiding principle of the transfer learning paradigm \cite{pan2009survey}. The last layers are retrained to suit a new domain, while the earlier layers containing the lower level representations are kept the same. \\
In the transformation from low-level feature information to a final class, a substantial amount of activation is suppressed by the network weights in the later layers and thereby `pruned away' in a sense. The feature maps in the middle layers of the CNN contain additional activation patterns which are irrelevant for inter-class classification, but could potentially be relevant for the intra-class classification of unique instances. Identifying signatures of traffic participants can be constructed from the mid-level feature maps of a sufficiently deep neural network. Deep neural features have previously been used to cluster images into semantic categories on which the networks producing these features were never explicitly trained \cite{donahue2014decaf}, which demonstrates the potential for generalization of deep features. \\
For this purpose, one relatively straightforward solution would be to "cut" all objects discovered by the YOLO network out of the scene image using the outputted bounding box coordinates, and feed the resulting images into a second pretrained deep image classification CNN, the final layers of which are slightly altered to extract features. This has previously been successfully done to measure the similarity of pedestrian detections for the purpose of accurate crowd counting \cite{yang2019pedestrian}. However, the addition of a second deep network would slow down the processing time of each frame considerably, which is a downside when aiming to identify vehicles and predict their trajectory in real-time in fast-paced traffic situations. \\

\subsubsection{Deep Integrated Feature Signatures}

\begin{figure}[b]
\centering
\includegraphics[width=\linewidth]{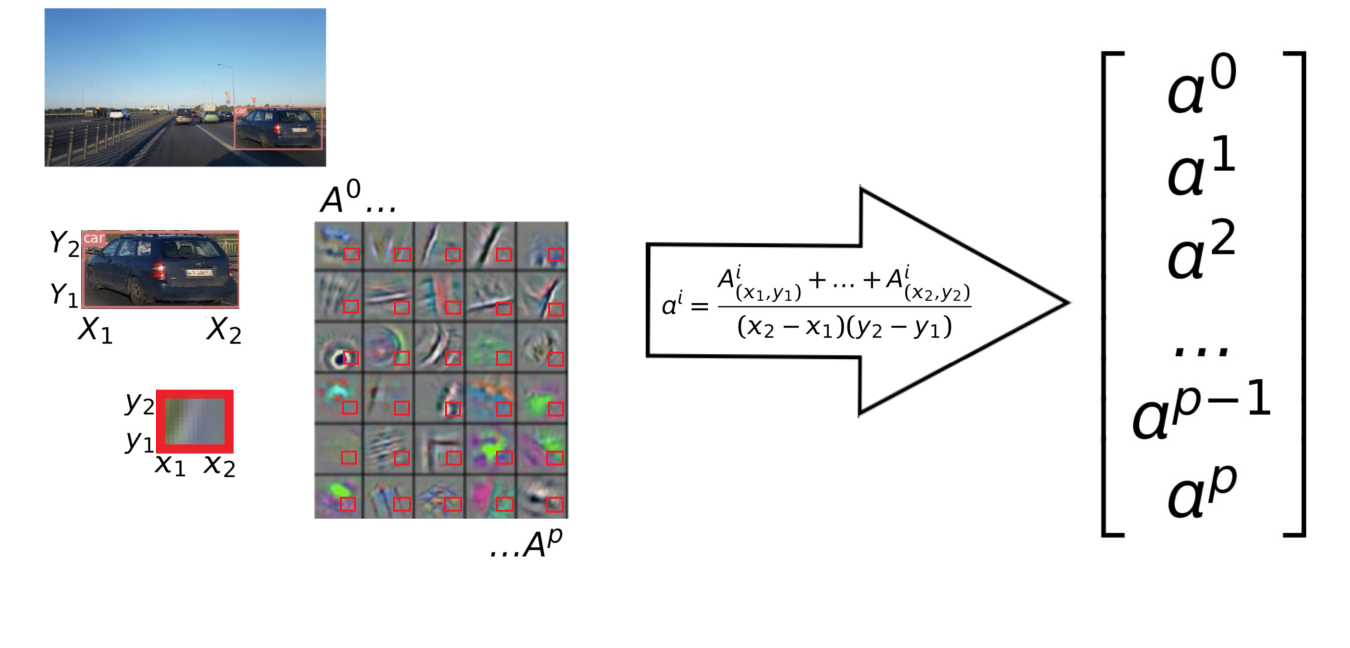}
\caption[DIFS Creation Process]{Diagram showing the process of DIFS from YOLO feature maps. Output bounding box coordinates $[X_1,X_2,Y_1,Y_2]$ of a detected object correspond to the relative region $[x_1,x_2,y_1,y_2]$ in $p$ feature maps at layer $A$ of the network, which are then used to create signature vector $a$ of length $p$ where $a^i$ is equal to the summed activation over region $[x_1,x_2,y_1,y_2]$ of feature map $A^i$.}
\label{fig:sig_creation}
\end{figure}

Ideally, the feature-based signatures used for instance classification should be extracted directly whenever an object is detected. In this research we demonstrate that the feature maps produced by an image detection network (YOLO v3) itself lend themselves directly to the creation of deep integrated feature signatures (DIFS) which can be used for to uniquely classify detected vehicle instances. When a vehicle is detected by the YOLO v3 network, we obtain the detection region from the bounding boxes of the network output. This region is then translated to the same relative region in feature map space for a predetermined layer of the network (referred to as the signature layer). Due to the fact that YOLO is fully convolutional, feature map information in this region was expected to accurately represent the detected object. \\
After determining the corresponding detection region in feature map space the summed activation over this region is subsequently calculated for each of the feature map images in the signature layer. This creates an n-dimensional signature for each vehicle detection, where n is the number of activation images in the signature layer. Fig. \ref{fig:sig_creation} shows the creation of a Deep Integrated Feature Signature from feature map information. \\
There is no trivial way of determining the optimal network layer to serve as signature layer. Therefore, performance of DIFS taken from the feature maps of signature layers at different stages in the YOLO network have been compared as part of previous research \cite{Oosterhuis2019who}. \\

\section{EXPERIMENTAL SETUP}
\label{setup}

\subsection{Datasets}

The instance classification performance of the deep integrated feature signatures is evaluated on two different datasets. In order to obtain a basis for possible comparison with other multi-vehicle tracking methods, the mono-camera image sequences of the KITTI benchmark data suite for autonomous driving is used \cite{Geiger2013IJRR}. Additionally the tracker performance is evaluated on previously unlabeled 4K dash-cam video data which was made available online \cite{majek2017object}. Data from this dataset has been annotated by hand for the purpose of this research. \\
The KITTI suite contains wide lens images with a 1242 by 373 pixel aspect ratio, and 30601 labeled vehicles. The manually annotated 4K dash-cam video dataset contains high definition images with a 3840 by 2160 pixel aspect ratio and 4317 labeled vehicle detections. The frame rate for the 4K video data is 30 frames per second, whereas the KITTI benchmark suite data was recorded at 10 frames per second.

\subsection{Experimental parameters}
To evaluate whether the Deep Integrated Feature Signatures obtained from the YOLO v3 network are an effective feature representation for the purpose of instance classification, we need to determine whether the intra-instance variations of DIFS are small enough compared to their inter-instance differences \cite{sun2014deep}. To assess the comparative intra-instance and inter-instance similarity of the DIFS a K nearest neighbor (KNN) classification was made, using five-fold cross validation. \\
Exploratory research revealed that both datasets are highly imbalanced with regard to instance frequency. The majority of the detections belong to only a small number of unique vehicle instances. This would bias instance classification using KNN towards those instances.\\
To counteract the skew in the datasets, a random sample of 20 examples for each unique vehicle with 20 or more occurrences was taken of each data set. Vehicles with less than 20 examples in the data were disregarded. This way two balanced sample sets were created. The KITTI sample set contains 340 samples (17 vehicle instances), whereas the 4K dash-cam sample set contains 700 samples (35 vehicle instances). \\

\section{RESULTS}
\label{results}

Table \ref{tab:results} shows the highest obtained classification accuracy of the DIFS instance classification evaluation for both datasets. This is 86.8\% classification accuracy on the KITTI sample set, and 96.7\% accuracy on the 4K dash-cam sample set. Both these results were obtained using DIFS obtained from layers in the middle of the YOLO v3 network (shortly before the `short cut' skip connection layers). These DIFS were classified using 1-nearest neighbor with a Manhattan distance measure. \\
Experimentally, we found the highest classification accuracy was obtained using KNN with $K=1$, compared to $K=3$ and $K=5$. It was also determined experimentally, that DIFS taken either from layers early in the network or from in the middle in the YOLO network performed better than DIFS taken from layers closer to the output layers. The classification accuracy did not differ noticeably when comparing the use of Manhattan and Euclidean distance measure in the KNN algorithm. For a detailed overview of these experiments and results, see \cite{Oosterhuis2019who}.

\begin{table}[t]
\centering
\caption{Accuracy and standard deviation of 5-fold 1NN instance classification using DIFS from the YOLO v3 network.}
\begin{tabular}{|l|l|l|}
\hline
dataset & accuracy & std. deviation \\ 
\hline
4K      & 96.7\%   & 1.2\%          \\
KITTI   & 86.8\%   & 2.1\%          \\
\hline
\end{tabular}

\label{tab:results}
\end{table}

\section{DISCUSSION}
\label{discussion}

When looking at the results from the evaluation of instance classification using 1-nearest neighbor on Deep Integrated Feature Signatures obtained from the YOLO v3 feature maps, we see that the highest obtained result is 96.7\% accuracy for the 4K dash-cam dataset and 86.8\% accuracy for the KITTI dataset (see Table \ref{tab:results}). In comparison, when using the SIFT key-point matching technique only 6.7\% classification accuracy was obtained on the 4K dash-cam video dataset, and only 5.9\% classification accuracy was obtained on the KITTI dataset \cite{Oosterhuis2019who}.  Our initial conclusion is that instance classification using DIFS from the YOLO feature maps highly successful, especially considering the fact that these feature maps are generated by the YOLO network during the detection pass, and the extra computational cost of generating the instance classification DIFS is therefore negligible. \\
Possible explanations for the difference in classification between the 4K dash-cam and KITTI datasets could be found in the difference in resolution, aspect ratio and frame rate between these two data sets. Due to the lower frame rate (10 frames/second) of the KITTI data, the similarity between images in consecutive time steps is lower than with the 4K dash-cam data (30 frames/second). Due to larger difference between width and height in the KITTI images, the images require more padding to fit the square sized requirement of YOLO, which reduces the area of relevant information of the KITTI images after preprocessing more than for the 4K dash-cam images. \\
An interesting result to note is that 1NN classification is the most successful and the KNN classification accuracy decreases with the number of neighbors used (See \cite{Oosterhuis2019who}). This indicates that the similarity of the DIFS is highly localized. This could likely be explained by the time dependence in the data. \\ 
It seems that with regard to the layer from which the signature is taken, the instance classification potential is high for a large part in the beginning and middle of the network, but decreases for layers deeper in the network (See \cite{Oosterhuis2019who}). A possible explanation for this is the fact that the layers used for the `last' DIFS are directly connected to the final YOLO output layers, which could mean that their activation is already highly class based. \\ 

\subsection{Future Work}

The pretrained YOLO v3 network was extraordinarly suited for instance classification of unique vehicles considering the fact that this network was trained on general object classes. Retraining the last network layers specifically on vehicle classes might improve the instance classification accuracy for this domain even further. \\
In this study three layers are used to create the deep integrated feature signatures and all DIFS are extracted from a single layer which is determined based on the output layer where the traffic participant is detected. This method of DIF-signature generation was developed experimentally and therefore remains an area of further experimentation. Instance classification based on the signature vectors was done with the KNN algorithm using Manhattan and Euclidean distance. A different classifier or similarity metric could possibly be used to equal or maybe even better effect. \\
Another interesting topic for future research would be to see if other deep neural architectures lend themselves generation of DIFS for instance classification just as well as the YOLO v3 architecture. Deep neural object detection is likely to move towards pixel-level classification \cite{redmon2018yolov3}, and the DIFS extraction method would need to be adapted to handle this change, as it currently relies on the bounding box to find the corresponding region in the feature maps. \\
In the context of situational awareness and multi-vehicle tracking in traffic, a pertinent topic of further research is the question of how the similarity of DIFS of objects in subsequent detection measurements could best be embedded into existing multi-vehicle detectors and trackers. One possible way would be to implement a DIFS similarity based regularization constraint similar to the enforced appearance similarity based on color histograms used in \cite{zamir2012global} and \cite{choi2015near}. \\

\section{CONCLUSION}
\label{conclusion}
In this study we sought to explore how the activity in the feature maps of a deep object detection network could be used for accurate real-time instance classification of unique vehicles, which could then help to further situational awareness of self-driving vehicles in combination with state-of-the-art multi-object detection and tracking techniques. \\
Instance classification of vehicles using Deep Integrated Feature Signatures obtained from the YOLO v3 network seems to be very successful. The highest results for instance classification were obtained when the signatures were created from feature maps in the beginning or the middle of the YOLO network and when using a 1-nearest neighbor classifier. The accuracy of instance classification using deep integrated feature signatures from YOLO feature maps also increased when the quality of the input video-stream data was higher. \\
The DIFS which were used for instance classification with high success were created from a pretrained YOLO v3 object detection network trained on general classes, which means the abstract representation of classes and objects of interest in the network may be generalizable accross different data domains. \\

In summary, deep integrated feature signatures created from the feature maps of the YOLO neural network trained for detection and multi-class classification have been shown to be very suitable for the purpose of instance classification. This provides us with a step towards realizing situational awareness of autonomous vehicles about surrounding participants in the dynamic environment of everyday traffic as well as potentially greatly improving existing multi-vehicle tracking systems. 

\addtolength{\textheight}{-12cm}   






\bibliographystyle{IEEEtran}
\bibliography{IEEEabrv,IEEEbibfile}

\begin{thebibliography}{10}
\providecommand{\url}[1]{#1}
\csname url@rmstyle\endcsname
\providecommand{\newblock}{\relax}
\providecommand{\bibinfo}[2]{#2}
\providecommand\BIBentrySTDinterwordspacing{\spaceskip=0pt\relax}
\providecommand\BIBentryALTinterwordstretchfactor{4}
\providecommand\BIBentryALTinterwordspacing{\spaceskip=\fontdimen2\font plus
\BIBentryALTinterwordstretchfactor\fontdimen3\font minus
  \fontdimen4\font\relax}
\providecommand\BIBforeignlanguage[2]{{%
\expandafter\ifx\csname l@#1\endcsname\relax
\typeout{** WARNING: IEEEtran.bst: No hyphenation pattern has been}%
\typeout{** loaded for the language `#1'. Using the pattern for}%
\typeout{** the default language instead.}%
\else
\language=\csname l@#1\endcsname
\fi
#2}}

\bibitem{Oosterhuis2019who}
T.~Oosterhuis, ``{`Who is Driving around Me?' Situational Awareness in Traffic
  using Deep Integrated Feature Signatures and Kalman Filtering.}'' Master's
  thesis, University of Groningen, 9700 AB Groningen, The Netherlands, 2019.

\bibitem{melo2004viewpoint}
J.~Melo, A.~Naftel, A.~Bernardino, and J.~Santos-Victor, ``Viewpoint
  independent detection of vehicle trajectories and lane geometry from
  uncalibrated traffic surveillance cameras,'' in \emph{Proc. International
  Conference on Image Analysis and Recognition (ICIAR)}.\hskip 1em plus 0.5em
  minus 0.4em\relax Springer, 2004, pp. 454--462.

\bibitem{danelljan2016beyond}
M.~Danelljan, A.~Robinson, F.~S. Khan, and M.~Felsberg, ``Beyond correlation
  filters: Learning continuous convolution operators for visual tracking,'' in
  \emph{Proc. European Conference on Computer Vision (ECCV)}.\hskip 1em plus
  0.5em minus 0.4em\relax Springer, 2016, pp. 472--488.

\bibitem{choi2015near}
W.~Choi, ``Near-online multi-target tracking with aggregated local flow
  descriptor,'' in \emph{Proc. IEEE International Conference on Computer Vision
  (ICCV)}, 2015, pp. 3029--3037.

\bibitem{scheidegger2018mono}
S.~Scheidegger, J.~Benjaminsson, E.~Rosenberg, A.~Krishnan, and
  K.~Granstr{\"o}m, ``Mono-camera 3d multi-object tracking using deep learning
  detections and pmbm filtering,'' in \emph{Proc. IEEE Intelligent Vehicles
  Symposium (IV)}.\hskip 1em plus 0.5em minus 0.4em\relax IEEE, 2018, pp.
  433--440.

\bibitem{zamir2012global}
A.~Zamir, A.~Dehghan, and M.~Shah, ``Gmcp-tracker: Global multi-object tracking
  using generalized minimum clique graphs,'' in \emph{Proc. European Conference
  on Computer Vision (ECCV)}, vol. 7573.\hskip 1em plus 0.5em minus 0.4em\relax
  Springer, 01 2012, pp. 343--356.

\bibitem{chan2016anticipating}
F.-H. Chan, Y.-T. Chen, Y.~Xiang, and M.~Sun, ``Anticipating accidents in
  dashcam videos,'' in \emph{Proc. Asian Conference on Computer Vision
  (ACCV)}.\hskip 1em plus 0.5em minus 0.4em\relax Springer, 2016, pp. 136--153.

\bibitem{pool2019context}
E.~A.~I. Pool, J.~F.~P. Kooij, and D.~M. Gavrila, ``Context-based cyclist path
  prediction using recurrent neural networks,'' in \emph{Proc. IEEE Intelligent
  Vehicles Symposium (IV)}, June 2019, pp. 824--830.

\bibitem{torstensson2019using}
M.~Torstensson., B.~Duran., and C.~Englund., ``Using recurrent neural networks
  for action and intention recognition of car drivers,'' in \emph{Proc.
  International Conference on Pattern Recognition Applications and Methods
  (ICPRAM)}, INSTICC.\hskip 1em plus 0.5em minus 0.4em\relax SciTePress, 2019,
  pp. 232--242.

\bibitem{frossard2018end}
D.~Frossard and R.~Urtasun, ``End-to-end learning of multi-sensor 3d tracking
  by detection,'' in \emph{Proc. IEEE International Conference on Robotics and
  Automation (ICRA)}.\hskip 1em plus 0.5em minus 0.4em\relax IEEE, 2018, pp.
  635--642.

\bibitem{roth2019deep}
M.~Roth, D.~Jargot, and D.~M. Gavrila, ``Deep end-to-end 3d person detection
  from camera and lidar,'' in \emph{Proc. IEEE Intelligent Transportation
  Systems Conference (ITSC)}, Oct 2019, pp. 521--527.

\bibitem{molchanov2017pedestrian}
V.~Molchanov, B.~Vishnyakov, Y.~Vizilter, O.~Vishnyakova, and V.~Knyaz,
  ``Pedestrian detection in video surveillance using fully convolutional {YOLO}
  neural network,'' in \emph{Automated Visual Inspection and Machine Vision
  {II}}, vol. 10334.\hskip 1em plus 0.5em minus 0.4em\relax International
  Society for Optics and Photonics (SPIE), 2017, pp. 193--199.

\bibitem{yang2019pedestrian}
X.~Yang., J.~Gaspar., W.~Ke., C.~T. Lam., Y.~Zheng., W.~H. Lou., and Y.~Wang.,
  ``Pedestrian similarity extraction to improve people counting accuracy,'' in
  \emph{Proc. International Conference on Pattern Recognition Applications and
  Methods (ICPRAM)}.\hskip 1em plus 0.5em minus 0.4em\relax {SciTePress}, 2019,
  pp. 548--555.

\bibitem{guidolini2018handling}
R.~Guidolini, L.~G. Scart, L.~F. Jesus, V.~B. Cardoso, C.~Badue, and
  T.~Oliveira-Santos, ``Handling pedestrians in crosswalks using deep neural
  networks in the {IARA} autonomous car,'' in \emph{Proc. International Joint
  Conference on Neural Networks (IJCNN)}.\hskip 1em plus 0.5em minus
  0.4em\relax IEEE, 2018, pp. 1--8.

\bibitem{he2017real}
X.~He and D.~Zeng, ``Real-time pedestrian warning system on highway using deep
  learning methods,'' in \emph{Proc. IEEE International Symposium on
  Intelligent Signal Processing and Communication Systems (ISPACS)}.\hskip 1em
  plus 0.5em minus 0.4em\relax IEEE, 2017, pp. 701--706.

\bibitem{redmon2018yolov3}
\BIBentryALTinterwordspacing
J.~Redmon and A.~Farhadi, ``{YOLOv3}: An incremental improvement,''
  \emph{Computing Research Repository (CoRR)}, vol. abs/1804.02767, 2018.
  [Online]. Available: \url{http://arxiv.org/abs/1804.02767}
\BIBentrySTDinterwordspacing

\bibitem{lin2014microsoft}
T.-Y. Lin, M.~Maire, S.~Belongie, J.~Hays, P.~Perona, D.~Ramanan,
  P.~Doll{\'a}r, and C.~L. Zitnick, ``Microsoft coco: Common objects in
  context,'' in \emph{Proc. European Conference on Computer Vision
  (ECCV)}.\hskip 1em plus 0.5em minus 0.4em\relax Springer, 2014, pp. 740--755.

\bibitem{redmon2016you}
\BIBentryALTinterwordspacing
J.~Redmon, S.~Divvala, R.~Girshick, and A.~Farhadi, ``You only look once:
  Unified, real-time object detection,'' in \emph{Proc. IEEE Conference on
  Computer Vision and Pattern Recognition (CVPR)}.\hskip 1em plus 0.5em minus
  0.4em\relax IEEE Computer Society, June 2016, pp. 779--788. [Online].
  Available: \url{https://doi.ieeecomputersociety.org/10.1109/CVPR.2016.91}
\BIBentrySTDinterwordspacing

\bibitem{lowe1999object}
D.~G. Lowe, ``Object recognition from local scale-invariant features,'' in
  \emph{Proc. International Conference on Computer Vision (ICCV)}.\hskip 1em
  plus 0.5em minus 0.4em\relax IEEE Computer Society, 1999, pp. 1150--1157.

\bibitem{pan2009survey}
S.~J. Pan and Q.~Yang, ``A survey on transfer learning,'' \emph{{IEEE} Trans.
  Knowledge Data Eng.}, vol.~22, no.~10, pp. 1345--1359, 2009.

\bibitem{donahue2014decaf}
J.~Donahue, Y.~Jia, O.~Vinyals, J.~Hoffman, N.~Zhang, E.~Tzeng, and T.~Darrell,
  ``Decaf: A deep convolutional activation feature for generic visual
  recognition,'' in \emph{Proc. International Conference on Machine Learning
  (ICML)}, 2014, pp. 647--655.

\bibitem{Geiger2013IJRR}
\BIBentryALTinterwordspacing
A.~Geiger, P.~Lenz, C.~Stiller, and R.~Urtasun, ``Vision meets robotics: The
  {KITTI} dataset,'' \emph{International Journal of Robotics Research (IJRR)},
  vol.~32, no.~11, pp. 1231--1237, 2013. [Online]. Available:
  \url{http://dblp.uni-trier.de/db/journals/ijrr/ijrr32.html\#GeigerLSU13}
\BIBentrySTDinterwordspacing

\bibitem{majek2017object}
\BIBentryALTinterwordspacing
K.~Majek, ``Object detection in 4k dashcam videos,'' 2017. [Online]. Available:
  \url{{https://medium.com/@karol\_majek/object-detection-in-4k-dashcam-videos-237c30ade356}}
\BIBentrySTDinterwordspacing

\bibitem{sun2014deep}
\BIBentryALTinterwordspacing
Y.~Sun, Y.~Chen, X.~Wang, and X.~Tang, ``Deep learning face representation by
  joint identification-verification,'' in \emph{Advances in Neural Information
  Processing Systems}.\hskip 1em plus 0.5em minus 0.4em\relax Curran
  Associates, Inc., 2014, pp. 1988--1996. [Online]. Available:
  \url{http://papers.nips.cc/paper/5416-deep-learning-face-representation-by-joint-identification-verification.pdf}
\BIBentrySTDinterwordspacing

\end{thebibliography}

\end{document}